\documentclass[10pt,twocolumn,letterpaper]{article}

\usepackage[pagenumbers]{wacv} % To force page numbers, e.g. for an arXiv version
\usepackage[accsupp]{axessibility} 
\usepackage{graphicx}
\usepackage{amsmath}
\usepackage{amssymb}
\usepackage{booktabs}
\usepackage{multirow}
\usepackage[pagebackref,breaklinks,colorlinks]{hyperref}

\usepackage[capitalize]{cleveref}
\crefname{section}{Sec.}{Secs.}
\Crefname{section}{Section}{Sections}
\Crefname{table}{Table}{Tables}
\crefname{table}{Tab.}{Tabs.}

\def\wacvPaperID{1442} % *** Enter the WACV Paper ID here
\def\confName{WACV}
\def\confYear{2024}

\begin{document}

\title{PathLDM: Text conditioned Latent Diffusion Model for Histopathology}

\newcommand*\samethanks[1][\value{footnote}]{\footnotemark[#1]}

\author{Srikar Yellapragada \thanks{Equal contribution. Correspondence to \href{mailto:myellapragad@cs.stonybrook.edu}{srikary@cs.stonybrook.edu}}
\and
Alexandros Graikos \samethanks
\and
Prateek Prasanna
\and
Tahsin Kurc
\and
Joel Saltz
\and
Dimitris Samaras\\\\
Stony Brook University\\
}

% \and

\maketitle

%%%%%%%%% ABSTRACT
\begin{abstract}
To achieve high-quality results, diffusion models must be trained on large datasets. This can be notably prohibitive for models in specialized domains, such as computational pathology. Conditioning on labeled data is known to help in data-efficient model training. Therefore, histopathology reports, which are rich in valuable clinical information, are an ideal choice as guidance for a histopathology generative model. In this paper, we introduce PathLDM, the first text-conditioned Latent Diffusion Model tailored for generating high-quality histopathology images. Leveraging the rich contextual information provided by pathology text reports, our approach fuses image and textual data to enhance the generation process. By utilizing GPT's capabilities to distill and summarize complex text reports, we establish an effective conditioning mechanism. Through strategic conditioning and necessary architectural enhancements, we achieved a SoTA FID score of 7.64 for text-to-image generation on the TCGA-BRCA dataset, significantly outperforming the closest text-conditioned competitor with FID 30.1. \footnote{Code is available at \href{https://github.com/cvlab-stonybrook/PathLDM}{https://github.com/cvlab-stonybrook/PathLDM}} 
\end{abstract}
%%%%%%%%% BODY TEXT
\section{Introduction}
Diffusion models have achieved impressive performance across diverse applications such as image generation, text-to-video synthesis, and audio generation \cite{sohl2015deep, ho2020denoising, ho2022imagen, kong2020diffwave}. Recently, Latent Diffusion Models (LDM) \cite{rombach2022high} have paved the way for higher-resolution text-conditioned image generation by employing a  two-step process involving image encoding and diffusion within the latent space. 

Unsurprisingly, diffusion models in histopathology image generation are an active area of research. Existing works such as \cite{moghadam2023morphology, xu2023vit, muller2023multimodal} are constrained by limited dataset size and conditioning capabilities. Our model performs a critical step toward the building of a \textit{foundational model}, which is to accommodate multiple modalities, including images, text, class labels, etc. 

% the current size of histopathology models is a few thousand images, which is not enough. existing models condition on labels. Due to recent advances in LLMs, embeddings produced by VLMs are often more expressive than manually picked labels. 
% don't say existing gap
% instead of enhancing the diffu

% previous pathology diffusion models couldn't achieve better results than a generic stable diffusion model because they were not trained with enough data. 

Conditioning on labeled data improves the diffusion model's generation quality by enabling data-efficient model training. As shown in \cite{nichol2021improved}, conditioning can be particularly effective in refining diffusion models, especially when dealing with complex data distributions. Existing histopathology diffusion models condition on class labels, devoid of textual input. Due to the recent advances in Vision Language models \cite{radford2021learning, Lu_2023_CVPR, huang2023leveraging}, embeddings produced by VLMs are often more expressive than manually picked labels. 

We introduce \textit{PathLDM}, the first Latent Diffusion Model for text-conditioned histopathology image generation. Our novel approach hinges on employing pathology text reports as the conditioning signal. Histopathology data is multimodal, with pathology reports often accompanying whole slide images. These reports often include detailed descriptions of cell types, disease classification, and other domain-specific insights pathologists provide.  They capture critical contextual information that is complementary to the diffusion model's generation ability.

We consider the extensively studied Cancer Genome Atlas Breast Invasive Carcinoma (TCGA-BRCA) dataset~\cite{abousamra2022deep, le2020utilizing}. Each entry within this dataset includes Whole Slide Images (WSI) and accompanying text reports provided by pathologists. Given that these reports can be lengthy (with a median length of 2800 characters) and unstructured, we adopt GPT's \cite{openaichatgptblog} capability to distill them into coherent summaries succinctly. These summaries serve as the guiding signal for PathLDM, effectively combining image and textual information. PathLDM produces visually coherent histopathology images aligned with the semantic details embedded within the text summaries through this synthesis.

Starting from the architecture of a Stable Diffusion \cite{rombach2022high} model, we propose a set of architectural modifications and develop PathLDM. Our technical modifications cover all three components of the LDM: the text encoder, the U-Net \cite{ronneberger2015u} denoiser, and the Variational Autoencoder (VAE). 

% Comparing previous models with SD, we noticed that fine-tuned SD performs better. This can be attributed to SD being trained on a much larger dataset. By conditioning with the appropriate pathology report text along with the necessary architectural changes, we're the first ones to show that a pathology diffusion model can beat stable diffusion. 

For the conditioning to work, we use a cyclical positional embedding in the text encoder. We pick a VAE that preserves the intricate details, such as cell structures and spatial layout of primitives, found in pathology images. We choose the appropriate U-Net, which acts as the right initialization for our VAE. We demonstrate that our effective conditioning, coupled with the architectural refinements, enable PathLDM to achieve a state-of-the-art Fréchet Inception Distance (FID) score of \textbf{7.64} for text-to-image generation on the TCGA-BRCA dataset. 

Comparing previous pathology diffusion models with Stable Diffusion, we observed that fine-tuned Stable Diffusion outperforms others. This can be attributed to Stable Diffusion being trained on a larger dataset. By conditioning with the appropriate pathology report text and the necessary architectural changes, we are the first to show that a pathology diffusion model can outperform fine-tuned Stable Diffusion. 
% We modify the architecture of PathLDM from that of the Stable Diffusion model \cite{rombach2022high} - a text-to-image Latent diffusion model trained on LAION-5B \cite{schuhmann2022laion}. Our enhancements cover all three components of the LDM: the Variational Autoencoder (VAE), the U-Net denoiser and the text encoder. We demonstrate that these refinements, coupled with strategic conditioning, enable PathLDM to obtain a remarkable state-of-the-art Fréchet Inception Distance (FID) score of \textbf{7.64} on the TCGA-BRCA dataset, specifically for text-to-image generation. Notably, this FID score represents a substantial 22-point improvement over the next best method.
% expand on the improvement

% a bespoke pathology diffusion model can achieve better results than stable diffusion

Our contributions are as follows:
\begin{itemize}
    \item We develop the first text-conditioned Latent diffusion model for histopathology image generation.
    \item We leverage GPT's capabilities to standardize and summarize pathology text reports for effective conditioning.
    \item We propose a framework that leverages conditioning with the appropriate architectural elements (cyclical positional embedding, VAE configuration, U-Net initialization) to improve the SoTA FID score dramatically. 
    \item We are the first to demonstrate that a bespoke Pathology diffusion model can perform significantly better than Stable Diffusion.
    
\end{itemize}

% We believe that our findings will lead to further research in domain-specific pathology diffusion models.

\section{Related Work}

The seminal work of diffusion models for image generation \cite{ho2020denoising} achieved comparable quality to ProgressiveGANs \cite{karras2017progressive} in unconditional CIFAR-10 \cite{krizhevsky2009learning} synthesis. Building on this, \cite{nichol2021improved} advanced DDPMs through class conditioning, learned variance and cosine noise scheduling, demonstrating competitive log-likelihoods in ImageNet \cite{russakovsky2015imagenet} generation. Similarly, \cite{dhariwal2021diffusion} improved the U-Net denoiser architecture with Adaptive Group Normalization, attention heads, and gradient-based sampling guidance, achieving SoTA results in ImageNet generation, rivaling BigGAN-deep \cite{brock2018large}.

% classfier free diffusion guidance [13] combines unconditional ....
% denoising diffusion implicit models [cite] accelerated diffusion sampling by ... 
% Latent diffusion models [35] bifurcate ...
Classifier-free diffusion guidance \cite{ho2022classifier} combines unconditional and conditional score estimates to achieve performance akin to Classifier guidance without training an auxiliary classifier. The acceleration of diffusion sampling by 10 to 50 times is proposed in \cite{song2020denoising} through the Denoising Diffusion Implicit Models (DDIM). Latent Diffusion Models, introduced in \cite{rombach2022high}, bifurcate the diffusion model training into two phases - an autoencoder for low-dimensional representation and a latent space diffusion model, thereby enhancing training efficiency. This approach surpasses previous methods in class conditional ImageNet generation and text conditional MS-COCO \cite{lin2014microsoft} synthesis \cite{nichol2021glide}. Our paper is the first to develop a text-conditioned Latent Diffusion Model for histopathology images.

The use of pretrained diffusion models for synthetic data creation has gained attention recently. In \cite{azizi2023synthetic}, the authors employ a text conditional model on a large scale to create synthetic data for ImageNet. Similarly, \cite{graikos2023conditional} leverage features extracted from an unconditional diffusion model's U-Net to provide guidance and enhance synthetic data quality. Both adopt the Classification Accuracy Score \cite{ravuri2019classification} as a crucial metric to evaluate the quality of synthetic data.

The inception of CLIP: Contrastive Language Image Pretraining \cite{radford2021learning} marked a significant advancement in Computer Vision pre-training. Their approach involved jointly training image and text encoders to develop a multi-modal embedding space. For text conditional diffusion models such as Stable Diffusion \cite{rombach2022high}, CLIP models were employed to encode text captions.

There have been recent advancements in Vision Language Models (VLMs) in the field of pathology. For instance, MI-Zero \cite{Lu_2023_CVPR} leverages a dataset comprising over 33k image-text pairs to train their VLM. Similarly, PLIP\cite{huang2023leveraging} gathers pathology data from Twitter to train a CLIP model. The authors of BioMedCLIP \cite{zhang2023large} introduce domain-specific adaptations tailored to biomedical VLP to outperform prior VLP approaches. However, none of these methods have utilized the pathology image-text pairs to train diffusion models.

The application of diffusion models in histopathology remains limited. The authors of \cite{moghadam2023morphology} train a pixel-level diffusion model for histopathology images, targeting 33,777 patches from low-grade gliomas within the TCGA dataset \cite{grossman2016toward}.  Similarly, \cite{xu2023vit} explore a Diffusion Autoencoder \cite{preechakul2022diffusion} trained on the NCT-CRC-HE \cite{kather} and PCam \cite{veeling2018rotation} datasets. Medfusion, introduced in \cite{muller2023multimodal}, is a Latent diffusion trained on 19,558 colorectal 
cancer images from CRCDX \cite{katherjakobnikolas_2020}.
%cancer images from the CRCDX \cite{katherjakobnikolas_2020} dataset.

Both \cite{moghadam2023morphology} and \cite{muller2023multimodal} focus on training class conditional models. Meanwhile, \cite{xu2023vit} opts for separate unconditional models per class, which is less efficient than a shared class conditional model. It's worth noting that these approaches employ relatively small datasets and lack a text conditioning mechanism. Our PathLDM cures both of these deficiencies.

% need a description for the figure
% the WSI and the accompanying pathology report ... 
\begin{figure}
    \centering
    \includegraphics[width=.95\linewidth]{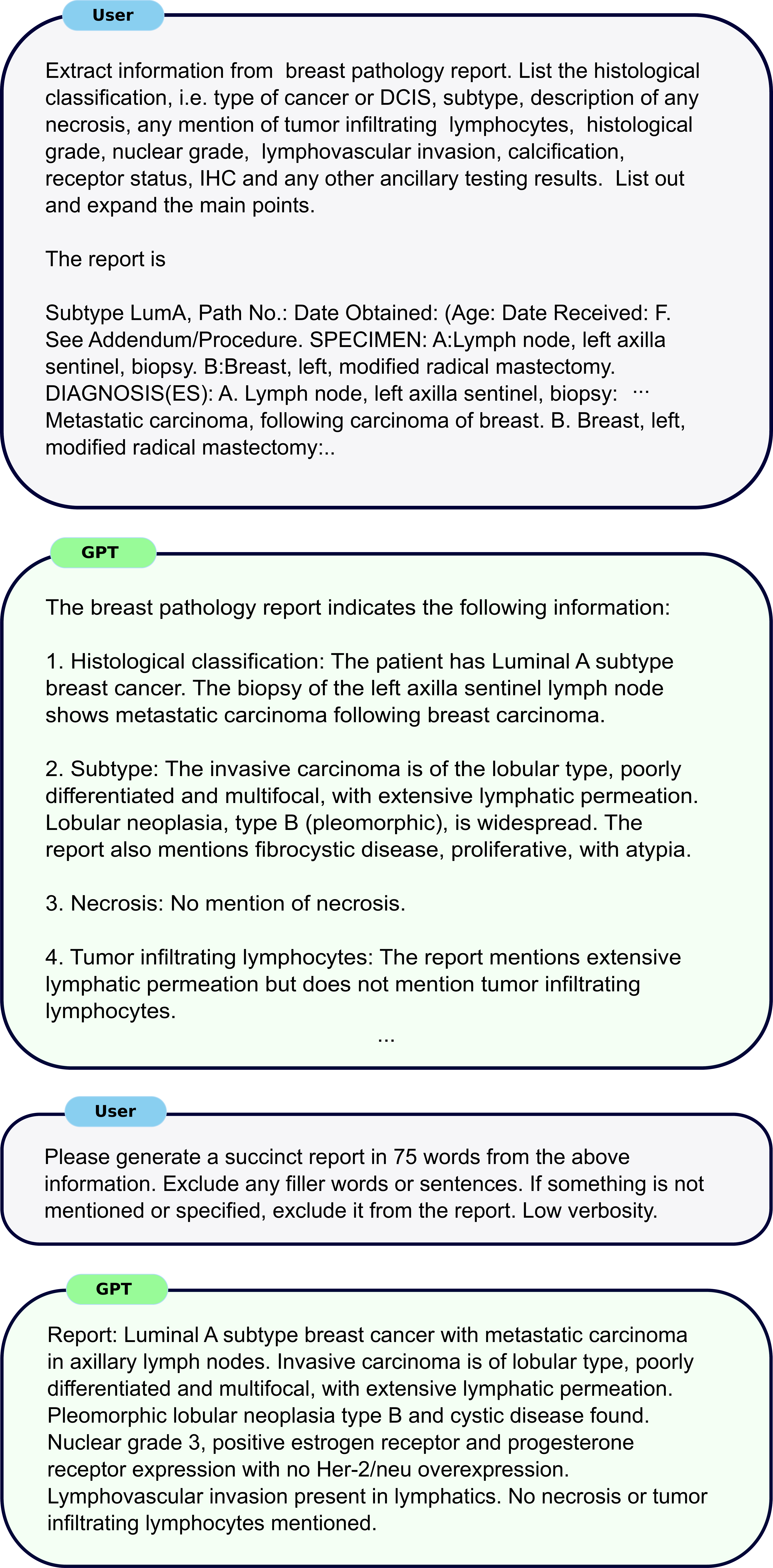}
    \caption{Example of our GPT-3.5 usage where we employ the chain-of-thought prompting technique to generate summaries. This approach leverages GPT’s capability to progressively develop a coherent line of reasoning, yielding a more refined and informative summary.}
    \vspace{-0.25cm}
    \label{fig:gpt}
\end{figure}
\begin{figure*}
    \includegraphics[width=\linewidth,trim={5pt 5pt 5pt 5pt},clip]{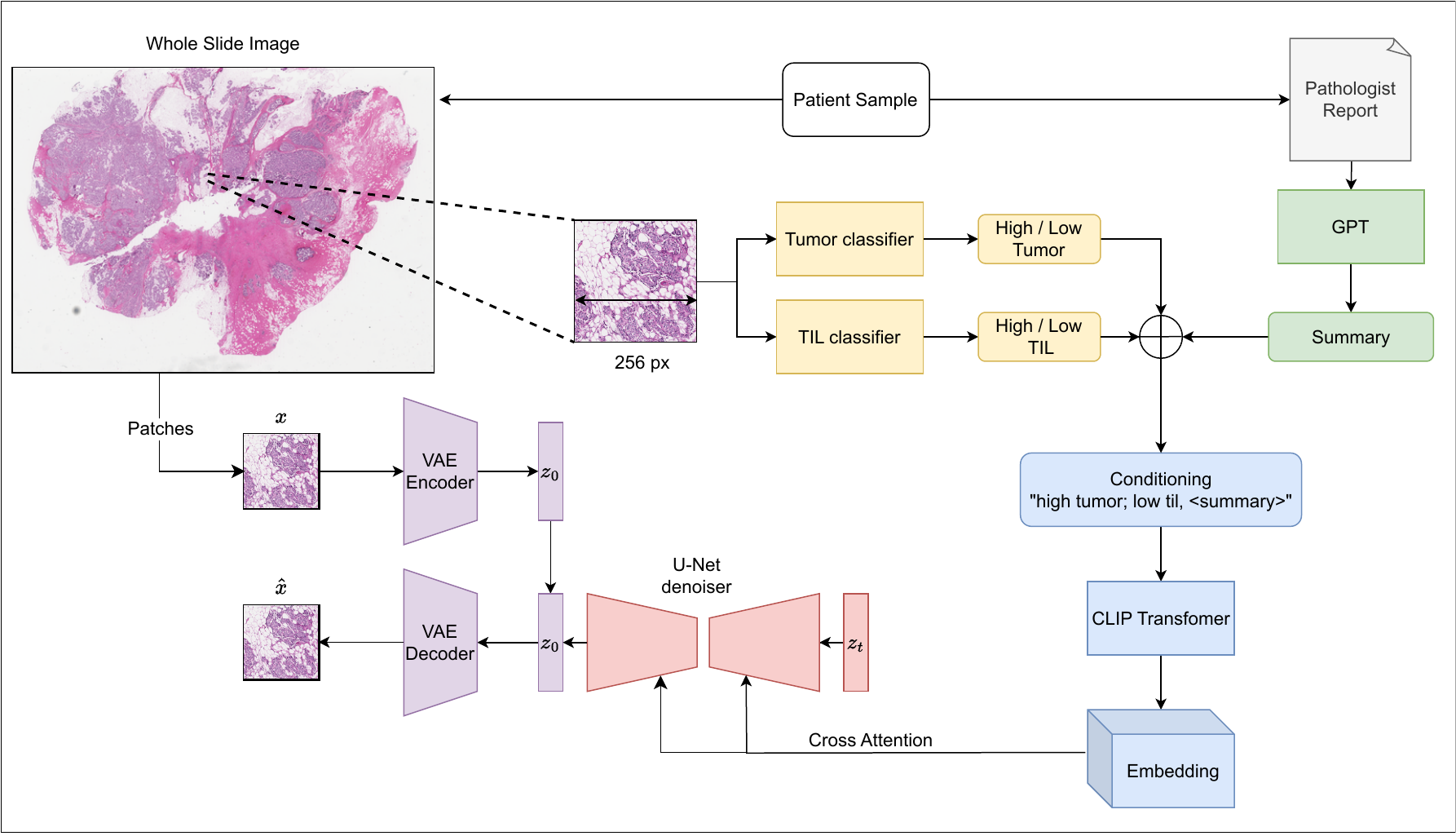}
     \caption{Overview of PathLDM. We start with a WSI and an accompanying Pathology report. We crop the WSI into 256 $\times$ 256 patches at 10x resolution. Leveraging GPT, we summarize the pathology report. For each patch, we compute Tumor and TIL probabilities, fuse them with the slide-level summary, and condition the LDM with CLIP embeddings of the fused summary. The VAE and text encoder remain frozen, and only the U-Net is trained. }
     \vspace{-0.25cm}
    \label{fig:overview}
\end{figure*}

\section{Methodology}
In this section, we describe our approach to conditioning and the necessary modifications applied to Latent diffusion models for the histopathology setting.

\subsection{GPT Summarization}
We leverage the TCGA-BRCA dataset for our model training. Each entry within this dataset includes a Whole Slide Image and an accompanying text report provided by pathologists, which we utilize as text conditioning. Accessible via cBioPortal \cite{cerami2012cbio, gao2013integrative}, these reports are primarily available in PDF format. Performing the complex process of Optical Character Recognition of PDFs to extract text is a nontrivial and challenging task. The authors of \cite{kefeli2023benchmark} tackled this challenge, extracting text for 9523 out of 11,000 TCGA text reports using OCR methods, further refined through manual validation. Their approach yielded a collection of extracted reports for 1034 BRCA WSIs, which we utilize.

The reports, on average, comprise around 2800 characters, roughly equivalent to 700 Byte-Pair Encoding tokens. These reports exhibit unstructured content due to the distinct writing style of each pathologist. SoTA text-to-image diffusion models, including Stable Diffusion, utilize a Transformer to encode the text captions, obtaining text embeddings. These embeddings are then cross-attended with image embeddings in the U-Net denoiser. Given that the cross-attention mechanism's computation complexity grows quadratically with sequence length, the need for concise summaries arises. Thus, we propose summarizing the reports and extracting crucial information using OpenAI's GPT model to ensure optimal processing within the model's limitations. It is worth noting that alternative LLMs, such as LLAMA \cite{touvron2023llama}, can also be considered for this purpose.

%Generative Pre-trained transformer (GPT), the prominent Large Language model developed by OpenAI, serves as a pivotal component in our approach for summarizing complex pathology text reports.
%It is built upon GPT-3.5 and fine-tuned for conversational applications using supervised and reinforcement learning techniques. 

We use the OpenAI GPT API to summarize the reports. Specifically, we opt for the \texttt{gpt-3.5-turbo} variant, built upon GPT-3.5 and designed for optimized chat interactions. %The authors of \cite{wei2022chain} introduce \textit{chain of thought prompting}. 
Zero-shot prompting is known to yield less consistent results than \textit{chain of thought prompting}. As described in \cite{wei2022chain}, this prompting technique generates a chain of thought through a series of intermediate reasoning steps and significantly improves the ability of Large Language models to perform complex reasoning.

% we randomly selected 5% and had them checked by board-certified pathologists.
% 

Following this idea, we also observed that breaking down the summarization process into multiple sequential calls improves summarization quality (as verified by our pathologists). Our methodology involves initially requesting GPT to outline the main points within the report. We then prompt the model to generate a concise summary based on the rehashed key information. This approach leverages GPT's capability to progressively develop a coherent line of reasoning, yielding a more refined and informative summary. We randomly selected $5\%$ of the generated summaries and had them validated by our board-certified pathologists. 

In our second user query, we prompt GPT to generate a report using 75 words, approximately 150 tokens. Due to the nature of GPT, this does not guarantee that the response will indeed remain within the specified limit. To address this, we also impose a  constraint (using the \textit{max\_tokens} parameter) on the final response generated by GPT, ensuring that it does not exceed 154 tokens. An example of an interaction that generates such a summary is presented in Fig.~\ref{fig:gpt}.

% \alex{Let's make this a figure instead in the interest of space. I'll put it together}
% \begin{itemize}
%     \item User Prompt 1: Extract information from breast pathology report. List the histological classification, i.e., type of cancer or DCIS, subtype, description of any necrosis, any mention of tumor-infiltrating lymphocytes,  histological grade, nuclear grade,  lymphovascular invasion, calcification, receptor status, IHC and any other ancillary testing results.  List out and expand the main points.

%     The report is \{ text report \}.

%     \item GPT: The breast pathology report indicates the following information: 
%     \SubItem{Classification: \{\}}
%     \SubItem{Tumor grade: \{\}} \ldots
%     \item User Prompt 2: Please generate a succinct report in 75 words from the above information. Exclude any filler words or sentences. If something is not mentioned or specified, exclude it from the report. Low verbosity.
%     \item GPT: A basal subtype infiltrative duct carcinoma mass (2.4 cm x 2.2cm x 2.2cm) with a high mitotic index \ldots 
% \end{itemize}

% method
% Use gpt to summarize, how we condition, 
% how we find that existing architecture fail
% do changes to make it work
% compare synthetic images with other medhodws, zoom in cells details
% 2d grid, rows are low to high til, cols are captions
% one for tumor
% linear probing
% cfg scale FID graph
% classification accuracy score for stable diffusion
% mention training reports are being used to generate synthetic images
% mention future work is generating synthetic reports
% generate medfusion synthetic data

\subsection{PathLDM}
The core of our methodology revolves around the development of PathLDM, a Latent Diffusion Model tailored for histopathology image generation with text conditioning. PathLDM's architecture consists of the Variational Autoencoder (VAE), the U-Net denoiser and the text encoder. We provide an overview of our method in Figure \ref{fig:overview}.

\subsubsection{Variational Autoencoder}
The VAE learns a latent representation from the input images. A significant criterion for evaluating the VAE's performance is its ability to reconstruct the original image through the decoder. Accurate reconstructions are crucial, as the subsequent diffusion process operates on these latent vectors and can, therefore, be constricted by the VAE's capabilities. 

Reconstruction quality becomes paramount when dealing with histopathology images that contain fine details such as cells. The authors of Medfusion \cite{muller2023multimodal} observed that using off-the-shelf Stable Diffusion's VAE can result in imperfect reconstructions. They modified the VAE architecture to overcome this issue by adjusting the number of channels and Convolution layer parameters. This refinement led to an increase in the Structural Similarity Index (SSIM) \cite{1284395} of the reconstructed images. 

However, both Stable Diffusion and Medfusion's VAE employ a downsampling factor of 8, effectively compressing 256 $\times$ 256 images down to a 32 $\times$ 32 $\times$ 4 (or 8 channels for Medfusion) latent. Opting for an overly small downsampling factor is undesirable, as it leads to a considerable increase in memory consumption and substantially hampers the training speed of the LDM. Thus, finding the right balance in compression levels is crucial for accurate reconstructions. 
% say what happens if downsampling factor is large

In PathLDM, we instead choose a 3-channel latent VAE with a downsampling factor of 4. In Section \ref{sec:expt_vae}, we delve into the impact of the downsampling factor on VAE's reconstruction quality when fine-tuned on Pathology images. We experimentally validate the advantages of our latent size choice by showing how we can maintain the fine structures found in large images. 
% To emphasize its significance, we extend our analysis to satellite imagery, a diverse context that includes smaller-sized objects, thereby demonstrating the broader applicability and implications of this aspect.
% \subsubsection{U-Net}

\subsubsection{U-Net Conditioning}

Leveraging GPT, we procure a single summary for each WSI. We then incorporate patch-level statistics corresponding to Tumor and Tumor-Infiltrating Lymphocyte (TIL) data to enrich these summaries. TILs have emerged as a promising biomarker in solid tumors, as a prognostic marker in triple-negative (TNBC) and HER2-positive breast cancer \cite{amgad2020report, denkert2010tumor, savas2016clinical}. We use the trained models from \cite{abousamra2022deep} and \cite{le2020utilizing} to obtain a Tumor and TIL probability value for each patch. These probability values were integrated into the text captions by first transforming them into an ordinal scale of Low - High and prepending them to the text description of each patch. This process created captions like ``High tumor; low til; \{GPT summary\}", effectively combining quantitative details with the summarization outcomes.

Our approach effectively fuses conditioning information derived from three distinct sources. By integrating the GPT summary, which encapsulates Whole-slide level conditioning, with Tumor and TIL statistics that represent patch-level information, we create an advantageous synthesis of global and local details. In contrast, existing Pathology diffusion models \cite{muller2023multimodal, moghadam2023morphology, xu2023vit} exclusively rely on class conditioning mechanisms, devoid of textual input. We delve into the impact of our conditioning mechanism in Section \ref{expt:text_cond}.

\begin{table*}[h]
\centering
\begin{tabular}{|c|c|c|c|c|c|c|}
\hline
VAE   & f & U-net training & \begin{tabular}[c]{@{}c@{}}Conditioning \\ network\end{tabular} & \begin{tabular}[c]{@{}c@{}}Conditioning\\ type\end{tabular} & \begin{tabular}[c]{@{}c@{}}Conditioning\\ modality\end{tabular} & FID $\downarrow$  \\ \hline\hline
vq-f8 & 8 & From scratch   & Class embedder                                                  & Tumor + TIL                                                 & Class label (4 classes)                                                       &   48.14    \\ 
vq-f4 & 4 & From scratch   & Class embedder                                                  & Tumor + TIL                                                 & Class label (4 classes)                                                       &   39.72    \\ 
vq-f4 & 4 & Finetune       & Class embedder                                                  & Tumor + TIL                                                 & Class label (4 classes)                                                       & 29.45 \\ 
vq-f4 & 4 & Finetune       & OpenAI CLIP                                                & Report + Tumor + TIL                                        & Text caption  (154 tokens)                                                  & 10.64 \\ 
vq-f4 & 4 & Finetune       & PLIP                                                            & Report + Tumor + TIL                                        & Text caption  (154 tokens)                                                  & \textbf{ 7.64}  \\ \hline
\end{tabular}
\caption{Effect of various components of our pipeline on FID. The use of text conditioning leads to the biggest reduction in FID.  f: VAE's downsampling factor.}
\vspace{-0.25cm}
\label{tab:fid-arch}
\end{table*}

\subsubsection{Encoding large prompts}

We use a CLIP \cite{radford2021learning} text encoder to process the text summaries obtained through GPT fused with Tumor and TIL presence. Its role is to transform these summaries into embeddings, compact yet rich textual context representations. Existing CLIP \cite{huang2023leveraging, radford2021learning} models are limited by a maximum context window length of 77 tokens. In our approach, we accommodate GPT summaries with a length of up to 154 tokens. This choice is deliberate, as it is twice the context length of the CLIP encoder. Opting for a length of 77 tokens would have resulted in an overly concise summary lacking in informative context.

To embed the full token sequence, we individually embed each of the halves and concatenate. This is equivalent to using a cyclical positional embedding that is repeated every 77 tokens, which we found to be sufficient for embedding the longer text sequences. We apply a causal mask and direct the full embedding sequence through the trained transformer encoder for processing. As an ablation, we replace the standard OpenAI CLIP \cite{radford2021learning} encoder with the more specialized PLIP encoder, discussed further in Section \ref{sec:expt_encoder}.

\section{Experiments}
\label{sec:expt}
In this section, we delve into the experimentation conducted to validate the effectiveness of our proposed PathLDM. We introduce architectural improvements to the Variational Autoencoder, U-Net denoiser, and text encoder components and explain how they affect the Fréchet Inception Distance (FID) metric.

We train all our diffusion models on 3 NVIDIA RTX 8000 GPUs with a batch size of 48 per GPU. We use the training code and checkpoints provided by \cite{rombach2022high}, implemented in Pytorch \cite{paszke2017automatic}. Our training employs a learning rate of $2\times10^{-5}$ with 10,000 warmup steps. For FID computation, we use the implementation from \cite{Seitzer2020FID}. 

We adopt DDIM sampling \cite{song2020denoising} with 50 steps and incorporate classifier-free guidance \cite{ho2022classifier} at a scale of 1.75 to generate synthetic images. We randomly sample classes for the class conditional models, and for the text conditional models, we randomly select summaries from the training set for conditioning.

\subsection{Dataset}
We selected 1136 Whole slide images from the TCGA-BRCA dataset and partitioned them into 977 for training and 199 for testing. Using the code from \cite{li2021dual}, we extract 256 $\times$ 256 patches at 10x magnification, culminating in 3.2 million training patches. Leveraging GPT, we procure a single summary for each WSI. 

We use the trained models from \cite{abousamra2022deep} and \cite{le2020utilizing} to obtain a Tumor and TIL probability value for each patch and incorporate them to enrich these summaries. This process created captions like ``High tumor; low til; \{GPT summary\}", effectively combining quantitative details with the summarization outcomes.

% We incorporated patch-level statistics corresponding to Tumor and Tumor-Infiltrating Lymphocyte (TIL) data to enrich these summaries. We use the trained models from \cite{abousamra2022deep} and \cite{le2020utilizing} to obtain a Tumor and TIL probability value for each patch. These probability values were integrated into the text captions by first transforming them into an ordinal scale of Low - High and prepending them to the text description of each patch. This process created captions like ``High tumor; low til; \{GPT summary\}", effectively combining quantitative details with the summarization outcomes.

\subsection{Baseline}
To assess the impact of different architectural changes on the FID, we begin with a class conditional LDM baseline. We exclusively utilize patch-level Tumor and TIL probabilities in this baseline without incorporating GPT's text summary. We first convert the probability values into an ordinal scale using a threshold of $0.5$, denoted as ``Low" and ``High." Subsequently, each patch is assigned a class label from a range of values: 0 representing ``Low Tumor + Low TIL", 1 for  ``Low Tumor + High TIL", 2 for ``High Tumor + Low TIL", and 3 for ``High Tumor + High TIL."

\subsection{VAE}
\label{sec:expt_vae}
To quantify the influence of VAE on the diffusion model's generation quality, we train two class-conditional LDMs, leveraging patch-level Tumor and TIL data. One model uses a VAE with a downsampling factor of 8, denoted as ``vq-f8", while the other employs a VAE with a downsampling factor of 4, denoted as ``vq-f4". As exhibited in Table \ref{tab:fid-arch}, FID shows noteworthy enhancement, improving from 48.14 to 39.72. This improvement underlines the substantial impact of VAE selection on the diffusion model's generative performance.

From the pretrained VAEs provided by \cite{rombach2022high}, we start with VAE VQ-f4 with a downsampling factor of 4. We then fine-tune it on our image patches. In Table \ref{tab:vae}, we highlight the significance of the downsampling factor or latent vector dimension in the quality of VAE reconstructions. Our choice of VAE yielded a significant boost in the SSIM of the reconstructed images, suggesting that overly aggressive compression may hinder the retrieval of intricate details, particularly in images featuring small cellular structures.

Our observations underscore the vital role of VAE selection, even in other large-image settings such as satellite imagery. The Deepglobe Landcover dataset \cite{Demir_2018} comprises 803 RGB satellite images, each measuring 2448x2448 pixels. We partition these images into 256x256 and 512x512 crops and reconstruct them using pre-trained VAEs sourced from \cite{rombach2022high}. The results presented in Table \ref{tab:vae-sat} provide compelling evidence that opting for a reduced downsampling factor substantially improves the SSIM of reconstructed images. This finding further highlights the significance of VAE choice in diverse image domains, including scenarios involving smaller-sized objects such as satellite imagery.

\begin{table}[]
\begin{tabular}{|c|c|c|c|c|}
\hline
VAE                                                                & Latent size & \begin{tabular}[c]{@{}c@{}}f\end{tabular} & SSIM  $\uparrow$         & MSE  $\downarrow$           \\ \hline\hline
\begin{tabular}[c]{@{}c@{}}Stable Diffusion \end{tabular} & 32x32x4     & 8                                                             & 0.874          & 25.795          \\ 
Medfusion                                                          & 32x32x8     & 8                                                             & 0.891          & 21.401          \\ 
\begin{tabular}[c]{@{}c@{}}Ours\end{tabular}             & 64x64x3     & 4                                                             & \textbf{0.961} & \textbf{11.503} \\ \hline
\end{tabular}
\caption{Comparison of reconstruction quality of VAE used in Stable Diffusion, medfusion and PathLDM, evaluated on TCGA-BRCA. The input image size is 256$\times$256$\times$3. A smaller downsampling factor (f) significantly boosts the SSIM. }
\label{tab:vae}
\end{table}

\begin{table}[]
\begin{tabular}{|c|c|c|c|c|c|}
\hline
VAE                               & \begin{tabular}[c]{@{}c@{}}Image\\ Resolution\end{tabular} & f                  & Latent size & SSIM $\uparrow$  & MSE  $\downarrow$ \\ \hline\hline
\multirow{2}{*}{VQ-f4}            & 256x256                                                & \multirow{2}{*}{4} & 64x64x3     & 0.958 & 2.359 \\  
                                  & 512x512                                                &                    & 128x128x3   & 0.957 & 2.246 \\ \hline
\multirow{2}{*}{VQ-f8} & 256x256                                                & \multirow{2}{*}{8} & 32x32x4     & 0.845 & 6.235 \\  
                                  & 512x512                                                &                    & 64x64x4     & 0.846 & 6.183 \\ \hline
\end{tabular}
\caption{Effect of latent size on the reconstruction quality of Satellite Images from Deepglobe \cite{Demir_2018} dataset. f: downsampling factor. The choice of VAE is significant even in diverse scenarios containing small objects.}
% \vspace{-0.3cm}
\label{tab:vae-sat}
\end{table}

\subsection{U-Net}

Moving forward, our next focus is the U-Net denoiser. To gauge the advantage of fine-tuning, we train two class conditional LDMs. In the first model, we initialize the U-Net denoiser weights from scratch. Second, we start with the ImageNet \cite{russakovsky2015imagenet} class conditioned U-Net denoiser weights from \cite{rombach2022high}. The outcome presented in Table \ref{tab:fid-arch} shows that employing a fine-tuned U-Net enhances the FID from 39.72 to 29.45. Intriguingly, even though the ImageNet weights are tailored to Natural images - distinct from the complexity of histopathology images - the learned denoising mechanism within the U-Net contributes significantly to the efficacy of our PathLDM training.

% our method avoids the artifacts that the other methods have
\begin{figure*}
    \centering
    \includegraphics[width=0.95\linewidth]{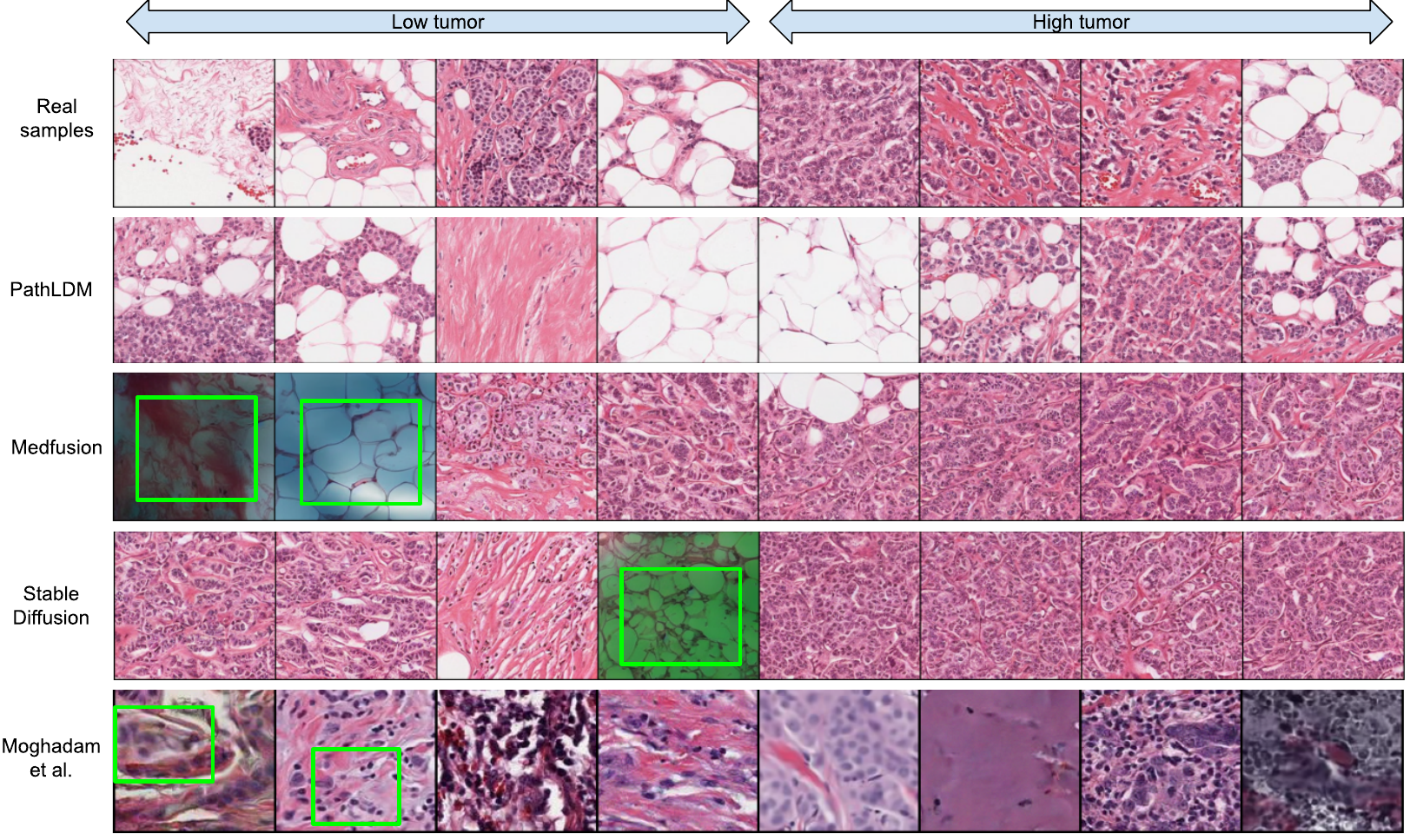}
    \caption{We choose a single text report and produce synthetic samples using Medfusion \cite{muller2023multimodal}, Stable Diffusion \cite{rombach2022high}, and PathLDM. Samples generated by Medfusion and Stable Diffusion show artifacts (indicated by green boxes) that are not present in our outputs. Moghadam et al. \cite{moghadam2023morphology} produces images in lower resolution ($128\times128$) and exhibits artifacts and blurriness (green boxes in the last row) in the output images. The first row contains the real samples from the corresponding WSI. }
    \vspace{-0.25cm}
    \label{fig:samples}
\end{figure*}

\subsection{Text conditioning}
\label{expt:text_cond}

To measure the concrete effects of our VAE and U-Net changes, we initially eshtablished a class conditional LDM baseline. Now, we aim to assess the benefit of introducing text summaries. Leveraging GPT, we obtain a summary for each Whole slide image (WSI). These summaries are combined with patch-level Tumor and TIL classes and employed to condition the diffusion model.

The results, as outlined in Table \ref{tab:fid-arch}, are impressive: the combination of text reports and patch-level statistics significantly reduces the FID score from 29.45 to 10.64. This improvement outperforms the gains achieved solely through architectural enhancements. Incorporating context from text summaries introduces a powerful interplay between visual and textual data, substantially enhancing model performance.  

\subsection{Text encoder}
\label{sec:expt_encoder}
Stable Diffusion uses the OpenAI CLIP \cite{radford2021learning} encoder, trained on Natural image caption pairs. As an ablation experiment, we introduce PLIP \cite{huang2023leveraging}, a CLIP model trained on pairs of pathology images and their corresponding text. As shown in Table \ref{tab:fid-arch}, adopting the PLIP encoder improved the FID from 10.64 to \textbf{7.64}. This improvement is attributed to the PLIP model's more adept grasp of the nuances present in pathology-specific text summaries.

\begin{table}[]
\begin{tabular}{|c|c|c|}
\hline
Method                     & Conditioning type & FID $\downarrow$                            \\ \hline \hline
Moghadam et.al \cite{moghadam2023morphology}             & Tumor + TIL                                                 & 105.81                         \\ \hline
\multirow{2}{*}{Medfusion \cite{muller2023multimodal}} & Tumor + TIL                                                 & 46.49                          \\
                           & Report + Tumor + TIL                                        & 39.56                          \\ \hline
Stable Diffusion \cite{rombach2022high}          & Report + Tumor + TIL                                        & 30.1                           \\ \hline
\multirow{2}{*}{PathLDM}      & Tumor + TIL                                                 & 29.45                          \\
                           & Report + Tumor + TIL                                        & \textbf{7.64} \\ \hline
\end{tabular}
\caption{Comparison of FID using 10k synthetic images. PathLDM outperforms other methods by a huge margin. Training Medfusion with text-conditioning improved the FID, highlighting the efficacy of our text summaries.}
\vspace{-0.5cm}
\label{tab:fid-methods}
\end{table}

\subsection{Comparison with other methods}
The cumulative architectural enhancements detailed in the preceding subsections result in the creation of our most potent Latent Diffusion Model - PathLDM. We compare our method against three alternative methods - Moghadam et al. \cite{moghadam2023morphology}, Medfusion  \cite{muller2023multimodal} and Stable Diffusion  \cite{rombach2022high}. 

For Moghadam et al. and Medfusion, both class conditional diffusion models, we train them with Tumor and TIL classes. Similarly, we train Stable Diffusion using text summaries, akin to our approach for PathLDM. We initialize the U-Net denoiser of Stable Diffusion and PathLDM using pretrained weights from \cite{rombach2022high}. However, Medfusion employs an 8-channel VAE, and no existing pretrained U-Net for Stable Diffusion has done more than four channels, which makes them unsuitable for initialization. Hence, we train it from scratch. 

The results, as shown in Table \ref{tab:fid-methods}, demonstrate the superiority of our text conditional PathLDM over other methods, establishing a significant FID gap. Furthermore, we also conducted text-conditional training for the Medfusion architecture, leading to an FID improvement from 46.49 to 39.56, highlighting the efficacy of our text summaries. Notably, even our class conditional model, trained without text reports, outperforms the class conditional variants of both Moghadam et al. and Medfusion, owing to the potency of the architectural enhancements we introduced.

Pixel-level diffusion models are computationally demanding, and no straightforward method exists to condition them on text captions. Recent works like GLIDE \cite{nichol2021glide} and Imagen \cite{ho2022imagen} are typically constrained to smaller image sizes (64 $\times$ 64) and rely on upsampling models to achieve high-resolution image synthesis. Since Moghadam et al. utilizes a pixel-level diffusion model, we haven't trained a text-conditional variant.   

For visualization, we select a text report and generate synthetic samples by employing Medfusion, Stable Diffusion, and PathLDM. Four samples each, categorized as "low tumor" and "high tumor," are generated for evaluation. The figures in Figure \ref{fig:samples} highlight the diversity and authenticity of the samples produced by our approach. Notably, our method generates samples that closely adhere to the distribution of actual samples. On the other hand, samples generated by Medfusion and Stable Diffusion show artifacts, such as blue stains, that are absent in our outputs.
% add examples of artifacts

\begin{figure*}[]
    \centering
    \includegraphics[width=\linewidth]{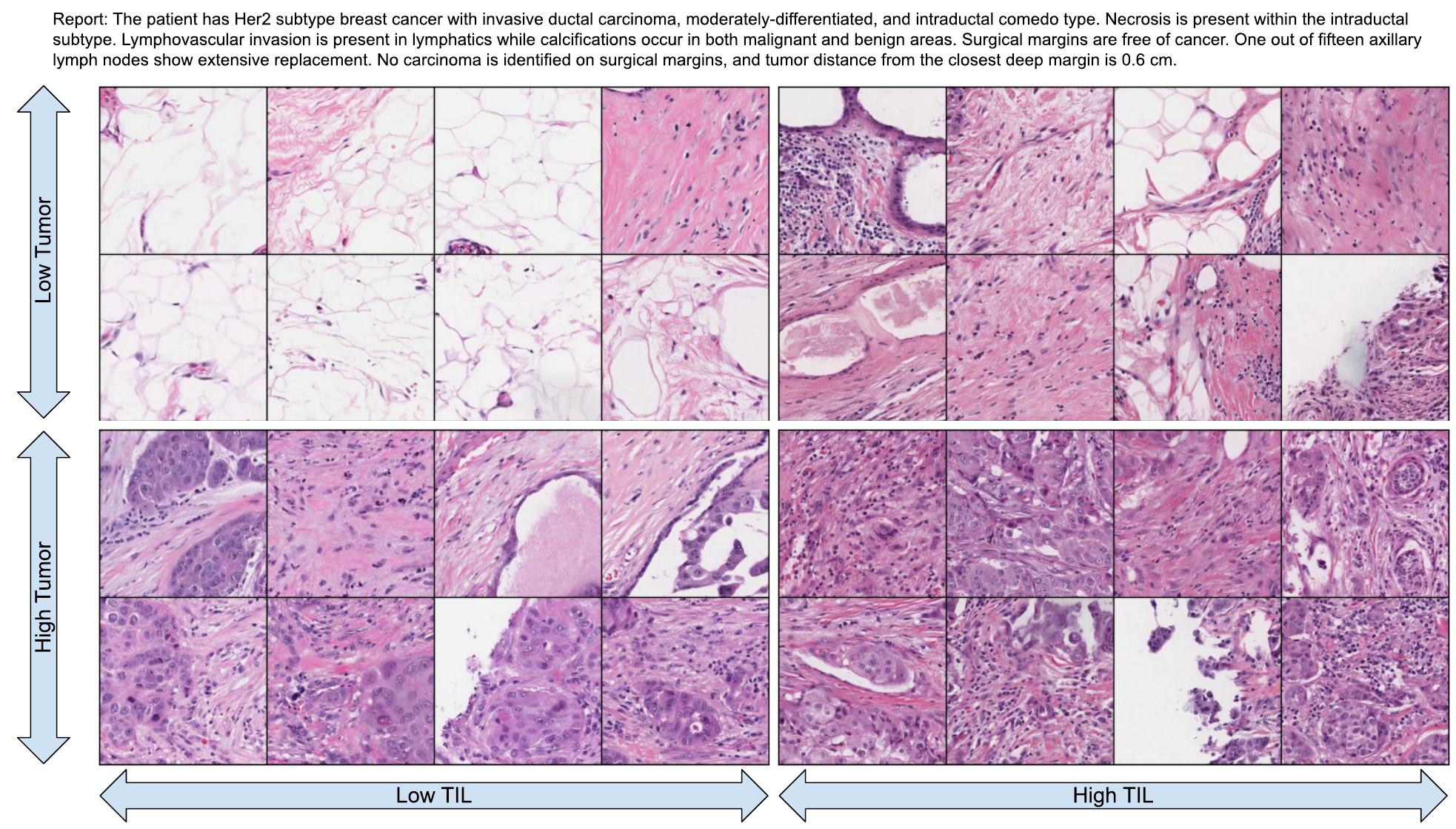}
    \caption{We present all combinations of low/high Tumor and low/high TIL samples generated by PathLDM. The corresponding GPT summary is provided on the top.}
    \label{fig_2}
\end{figure*}

\begin{table}[]
\centering
\begin{tabular}{|c|c|c|}
\hline
Data type & Data source & Accuracy (\%) \\ \hline \hline
Real      & TCGA-BRCA   & 90.31         \\ \hline
Only Synthetic & Medfusion   & 79.12          \\ 
Only Synthetic & PathLDM        & \textbf{83.32}         \\ \hline
Real + Synthetic & PathLDM        & \textbf{90.74}         \\ \hline
\end{tabular}
\caption{Classification Accuracy scores of a ResNet-34 on synthetic data. Higher CAS proves the effectiveness of PathLDM-generated synthetic samples in capturing significant features. }
\vspace{-0.15cm}
\label{tab:cas}
\end{table}

\subsection{Classification Accuracy Score}
In addition to the widely used FID for evaluating generative models, an alternative evaluation metric is the Classification Accuracy Score (CAS) \cite{ravuri2019classification}. To compute CAS, a classifier is trained on synthetic data and evaluated on real validation data. To train PathLDM, we use a dataset of 3 million patches, each associated with a Tumor label of either ``Low" or ``High." We create synthetic datasets using PathLDM and Medfusion, generating 3 million patches. We assign the Tumor label used in generating each patch as its ground truth. 

We then train ImageNet pretrained ResNet-34 \cite{he2016deep} classifiers on these synthetic datasets for 40 epochs with a batch size 256 and the Adam optimizer \cite{kingma2014adam}. We use a learning rate of $10^{-3}$ for the first 20 epochs, then a reduction to $\frac{1}{10}$th at epochs 20 and 30. As presented in Table \ref{tab:cas}, the classifier trained on our synthetic data achieves a validation accuracy of \textbf{83.32\%}, significantly higher than a classifier trained on Medfusion's data (79.12\%). When the classifier is trained using a mix of real and synthetic data, there is a modest improvement in accuracy, increasing from 90.31\% to \textbf{90.74\%}. This marginal enhancement suggests that the task of tumor classification is inherently less complex.

% \begin{figure}
%     \centering
%     \includesvg[width=0.5\linewidth]{fig/sweep_guidance_str.svg}
%     \caption{Training set FID vs. classifier-free guidance weight.}
%     \label{fig:guidance_str}
% \end{figure}

\section{Ablation studies}

\subsection{Sampling parameters}
We adopt DDIM sampling with 50 steps and integrate classifier-free guidance. We conduct a parameter sweep across guidance values ranging from $0.1$ to $3$, resulting in the optimal FID outcome at $1.75$. This value is then employed consistently across all our experimental scenarios.

\begin{table}[]
\centering
\begin{tabular}{|c|c|}
\hline
PathLDM Variant             & FID   \\ \hline\hline
Baseline - True summary            & 7.64  \\ 
Random summary   & 22.16 \\ 
5 summaries per WSI & 7.34  \\ \hline
\end{tabular}
\caption{Ablation to highlight the benefit of text summaries.}
% \vspace{-0.15cm}
\label{tab:text-abl}
\end{table}

\subsection{Benefit of text reports}
We perform two ablation experiments to underscore the significance of text report conditioning. In the first ablation, we train an LDM where, for each patch, we substitute its Whole Slide Image (WSI) summary with a randomly chosen summary from a different WSI. In the second ablation, we extend the utility of GPT by generating five distinct summaries from each text report obtained through five separate GPT API queries. Each of these queries operates independently and adheres to the same token limitations. Subsequently, for each patch, we randomly select only one summary from the five provided by GPT.

The outcomes, as shown in Table \ref{tab:text-abl}, provide valuable insights - substituting a random summary leads to the deterioration of the diffusion model, increasing the FID from 7.64 to 22.16. This outcome demonstrates that using an inappropriate summary or the absence of one altogether worsens the generation quality. Conversely, deploying five summaries per WSI improved the FID from 7.64 to \textbf{7.34}. These ablations underline the crucial role that accurate text conditioning plays in enhancing the performance of our model. 

% \section{Limitations}
% A limitation of our approach lies in the re-use of text prompts from the training set for synthetic image generation. A promising avenue for future exploration would be maximizing the potential of GPT by generating novel pathology reports based on the dataset of existing ones. Employing these fresh summaries for image generation could further reveal the model's capabilities in controlling the visual features from text descriptions.

% say we haven't use held out reports for generating images. it's a limitation and a future work. 

\section{Limitation}
A limitation of our approach is the reliance on text prompts from the training set for generating synthetic images. We did not employ held-out reports for image generation, which could be a potential avenue for future research.

\section{Conclusion}

Our paper presents PathLDM, a groundbreaking Latent Diffusion Model designed for generating histopathology images conditioned on text.  By effectively summarizing complex pathology text reports using GPT, we establish a robust conditioning mechanism that bridges the gap between textual and visual content. Our strategic conditioning, coupled with architectural enhancements, led to a SoTA FID score of 7.76 on the TCGA-BRCA dataset, significantly outperforming the closest competitor with FID 30.1. This study performs a critical step toward the building of a foundational model, which is accommodating multiple modalities, including images, text, class labels, etc. We anticipate that our work will encourage future exploration of domain-specific pathology diffusion models.

\section*{Acknowledgements}
The reported research was partially supported by NCI awards 5U24CA215109,
1R21CA258493-01A1, NSF grants IIS-2123920, IIS-2212046 and Stony Brook Profund 2022 seed funding.

%%%%%%%%% REFERENCES
{\small
\bibliographystyle{ieee_fullname}
\bibliography{references}
}

\end{document}